\documentclass[runningheads]{llncs}
\usepackage{amsfonts}
\usepackage{amsmath}
\usepackage{booktabs} 
\usepackage{graphicx}
\usepackage{hyperref} 
\usepackage{algorithm}
\usepackage{algorithmic}

\usepackage{algorithm}
\usepackage{booktabs}
\usepackage[table]{xcolor} 
\usepackage{multirow} 

\usepackage{marvosym}
\begin{document}
\title{Algorithmic Identification of Essential
Exogenous Nodes for Causal Sufficiency in Brain
Networks}
%
%
\author{
Abdolmahdi Bagheri\inst{1}$^*${(\Letter)}  \and Mahdi Dehshiri\inst{1}$^*$
\and
Babak Nadjar Araabi\inst{1}\and
Alireza Akhondi-Asl\inst{2}}
\authorrunning{A. Bagheri et al.}
%
\institute{School of Electrical and Computer Engineering, University of Tehran, College of Engineering, Tehran, Iran \\
\email{abdolmahdibagheri@ut.ac.ir}
\and
Department of Anaesthesia, Harvard Medical School, Boston, Massachusetts, USA\\
\def\thefootnote{*}\footnotetext{Abdolmahdi Bagheri and Mahdi Dehshiri contributed equally.}\def\thefootnote{\arabic{footnote}}
}
\maketitle              
\begin{abstract}
In the investigation of any causal mechanisms, such as the brain's causal networks, the assumption of causal sufficiency plays a critical role. Notably, neglecting this assumption can result in significant errors, a fact that is often disregarded in the causal analysis of brain networks.
In this study, we propose an algorithmic identification approach for determining essential exogenous nodes that satisfy the critical need for causal sufficiency to adhere to it in such inquiries. Our approach consists of three main steps: First, by capturing the essence of the Peter-Clark (PC) algorithm, we conduct independence tests for pairs of regions within a network, as well as for the same pairs conditioned on nodes from other networks. Next, we distinguish candidate confounders by analyzing the differences between the conditional and unconditional results, using the Kolmogorov-Smirnov test.
Subsequently, we utilize Non-Factorized identifiable Variational Autoencoders (NF-iVAE) along with the Correlation Coefficient index (CCI) metric to identify the confounding variables within these candidate nodes. Applying our method to the Human Connectome Project’s (HCP) movie-watching task data, we demonstrate that while interactions exist between dorsal and ventral regions, only dorsal regions serve as confounders for the visual networks, and vice versa. These findings align consistently with those resulting from the neuroscientific perspective. Finally, we show the reliability of our results by testing 30 independent runs for NF-iVAE initialization.

\end{abstract}
\section{Introduction}

Probing a causal model and extracting its links necessitates upholding certain prerequisites, among them, the causal sufficiency assumption is the primary one \cite{spirtes2000causation}. With various definitions for this assumption presented in \cite{beckers2021causal}, it essentially asserts that in exploring causal relations, all potential confounders or their ancestors must be accounted for the model under investigation. Violating this assumption often results in extracting spurious connections, suggesting causal links where none exist. 
Despite the importance of this assumption, it is often overlooked in studying causality in brain networks. In these studies, some focused on only one network such as the visual, working memory, or attention network, as exemplified by those in \cite{nauhaus2009stimulus,harris2011cortical,hofer2011differential,koshino2005functional,hampson2006brain}, or two or more networks together such as those in \cite{ungerleider2000mechanisms,luo2015neuronal}, working memory and attention networks in \cite{oberauer2019working,panichello2021shared,awh2006interactions}, dorsal and ventral attention networks in \cite{vossel2014dorsal,suo2021anatomical,vossel2012deconstructing} and the dorsal attention, ventral attention, and visual presented in \cite{zhao2022involvement,ahrens2019both,majerus2012attention}.
However, in many cases, the confounding variables can be out of the considered networks, for instance, in the finger-tapping task, the information flows from the brainstem implying that subcortical regions can act as confounding variables for the cortex, highlighting that these regions must be included in extracting causality to uphold this assumption.
While studies such as those by \cite{bagheri2023brain,bagheri2023bayesian} have made strides by incorporating subcortical regions alongside the cortex in causal discovery, leveraging a fusion of methodologies outlined in \cite{zheng2018dags,ng2020role,bello2022dagma,andrews2024fast}, their findings still suffer from unreliability due to the low performance of existing causal discovery methods in extracting causality in large networks. Conversely, the approaches presented in \cite{forre2018constraint,gerhardus2020high,kaltenpoth2023nonlinear,wang2023causal} have demonstrated progress in addressing causal discovery in the presence of both measured and unmeasured confounders, enabling the study of smaller networks without the concern of causal sufficiency. However, they also suffer from low performance in real-world scenarios. This suggests that focusing solely on a small network without causal sufficiency may not be practical with the existing methods. An alternative approach could involve identifying and exclusively including confounders in the network under study, followed by employing existing methods that extract causality with high accuracy in smaller networks.

In this paper, we present an algorithmic identification approach to extract the essential exogenous nodes for causal sufficiency in brain Networks. This enables the use of existing causal methods that accurately extract causal structures for smaller networks by including only the confounding variables, thus mitigating concerns about the low performance of existing methods with a large number of variables. 
Our proposed approach performs in three sequential steps. Firstly, similar to the Peter-Clark (PC) algorithm \cite{spirtes2000causation}, initiates by conducting unconditional and conditional independence tests between pairs of regions within a network. Then, it carries out conditional dependency tests on these same pairs conditioned on regions from other networks. This procedure is uniformly applied across all participants. By comparing the distribution of conditional and unconditional independence tests for all subjects with the Kolmogorov-Smirnov test, the candidate confounders are distinguished. Finally, 
the results from applying Non-Factorized Identifiable Variational Autoencoders (NF-iVAE) \cite{lu2021invariant} to the candidate regions are computed and based on the Correlation Coefficient index (CCI) metric, we identify confounding variables among the candidates.
The proposed algorithm is population-based, implying that it is applicable in problems involving a cohort of individuals, such as brain network analysis.

Our contributions are threefold:
\begin{itemize}
    \item We introduce an algorithmic identification method for discovering essential exogenous nodes for causal sufficiency.
    \item We apply our approach to the Human Connectome Project’s (HCP) movie-watching task data and demonstrate that the identified regions align consistently with those discovered from a neuroscientific perspective.
    
    \item we show the reliability of our results by testing 30 independent runs for NF-iVAE initialization. 
\end{itemize}
The codes associated with this study are publicly available at \href{https://github.com/mhdehshiri/Algorithmic-Identification-of-Essential-Exogenous-Nodes-for-Causal-Sufficiency-in-Brain-Network}{github}.

\section{Proposed method}

To present and discuss the design of the proposed identification method, we first present definitions and preliminaries. Then the algorithm for identifying essential exogenous nodes for causal efficiency is presented. Finally, we describe the data used and the preprocessing steps.
In this paper, $v \in V$ shows all the variables. $z \in Z$ are the variables in the network under the study. $s \in S$ is variables out of the networks under study, i.e., candidate confounders and $c \in C$ are the confounders. $j \in J$ denotes the index of the subjects in the set $J$. 
\subsection{Definitions and preliminaries}

\subsubsection{Causal sufficiency}\
A general definition of causal sufficiency assumption is presented in \cite{beckers2021causal} as follows,

\begin{definition}[A general definition of causal sufficiency]
The set of variables in $S\subseteq V$ is causally sufficient for all the variables in $Z\subseteq V$, if, for every pair of variable $z_1$ and $z_2$, all the common causes or at least the ancestor of common causes are in $S$, i.e., $C\subseteq S\subseteq V$.
\end{definition}
The details on the PC algorithm, an (un)conditional independence test, and the NF-iVAE are presented in supplementary materials.

\subsection{Identification of Essential exogenous nodes for causal sufficiency}

Our proposed identification algorithm runs in three main steps:

\begin{enumerate}
     \item Our approach, similar to the PC algorithm, initiates by conducting unconditional and conditional independence tests between all pairs of regions, $Z$. Complementary to the PC algorithm, it then carries out conditional independence tests on these same pairs conditioned on regions from other networks, $S$ is the number of possible confounders. This procedure is uniformly applied across all participants.
     \item By comparing the distribution of conditional and unconditional independence tests of each pair for all subjects with the Kolmogorov-Smirnov (KS) test, a subset of candidate confounding variables are distinguished. These candidates can represent both a confounder and a mediator.

     \item Measuring the Correlation coefficient index (CCI) between the latent variables generated by NF-iVAE and the real values of candidates proposed in the previous step, we identify the nodes that are confounding variables of the network under study.
\end{enumerate}

The proposed 3-step approach for the identification of confounders is given in Algorithm \ref{alg:cap}.
\begin{algorithm}
\textbf{Input:} Subset of Nodes $Z \subset V$, Subset of potential confounding variables $L \subset V \setminus Z$, List of subjects $J$, Correlation Coefficient Threshold $T_h$, $\alpha$: Threshold for $H_0$. \\
\textbf{Initialize:} Confounding candidate variables set $S \leftarrow \{\}$, Confounding variables set $C \leftarrow \{\}$.\\
\textbf{Step 1:}
\begin{algorithmic}[1]
\FOR{$l \in L$}
    \FOR{$ z_i \in Z$}
        \FOR{$z_k \in Z \setminus \{z_i\}$}
            \STATE Initialize unconditional independence set $U_{\text{unc}} \leftarrow \{\}$
            \STATE Initialize conditional independence set $U_{\text{cond}} \leftarrow \{\}$
            \FOR{$j \in J$}
                \STATE $U_{\text{unc}} \leftarrow U_{\text{unc}} \cup \{z_i^j \perp z_k^j\}$
                \STATE $U_{\text{cond}} \leftarrow U_{\text{cond}} \cup \{z_i^j \perp z_k^j \mid l^j\}$
            \ENDFOR \\
            \textbf{Step 2:}
            \STATE $pval \leftarrow \text{\{PValue of KS-test on } U_{\text{cond}} \text{ and } U_{\text{unc}}\}$ 
            \IF{$pval < \alpha$ \AND mean($U_{\text{unc}}$) $<$ mean($U_{\text{con}}$)}
                \STATE $S \leftarrow S \cup \{l\}$
            \ENDIF \\
            \textbf{end Step2}
        \ENDFOR
    \ENDFOR
\ENDFOR
\\
\textbf{end Step1}
\end{algorithmic}
\textbf{Step 3:}
\begin{algorithmic}[1]
  \STATE Apply NF-iVAE on $Z$ and estimate $L'$.
  \STATE $CCI \leftarrow \text{\{Correlation Coefficient Index between $L$ and $L'$ pairs\}}$
  \FOR{$ s \in S$}
    \IF{$CCI[s] > T_h$}
    \STATE $C \leftarrow C \cup \{s\}$
    \ENDIF
  \ENDFOR
\end{algorithmic}
\textbf{end Step3}\\
 {\textbf{Output:}
  $ C $.}
 \caption{Identification of Essential Exogenous Nodes for Causal Sufficiency}\label{alg:cap}
\end{algorithm}
\begin{figure}

\includegraphics[width=\textwidth]{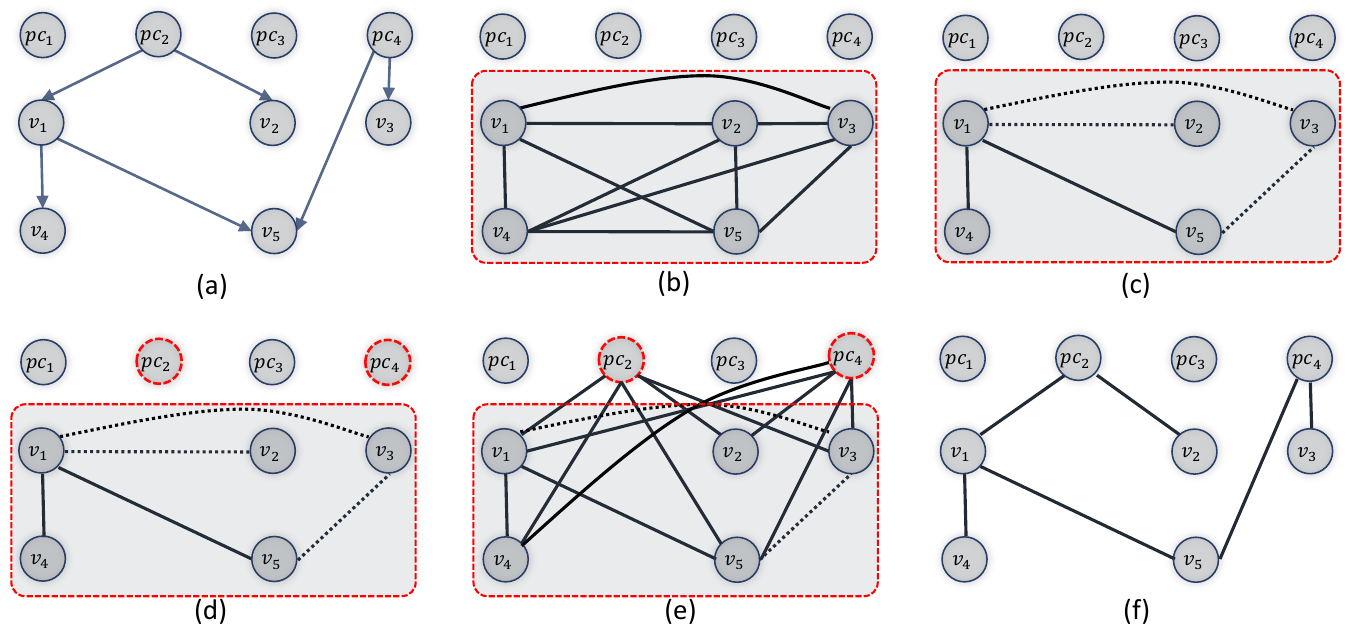}
\caption{a.The true underlying network with variables $z_1, ..., z_5$, and confounders, $c_1$ and $c_2$ b. starting PC algorithm with fully connected graph, c. result of PC algorithm applying d. Identifying essential confounders based on the proposed algorithm i.e., $s_2=c_1$ and $s_4=c_2$, e. applying PC algorithm to $z_1,..., z_5$ and $c_1$ and $c_2$, F. result of applying causal discovery method with confounders}\label{fig2}
\end{figure}

Fig. \ref{fig2} illustrates the result of employing algorithmic identification of confounding variables. In this figure, the true graph is presented in (a). In (b), the set of variables within the dashed rectangle represents the network under study. In (c), the result of the PC algorithm is shown, which includes spurious connections. In (d), the essential confounders are extracted. In (e), all possible links from the confounders and variables of the network under study are added. Finally, in (f), the result of applying PC to this set of variables is illustrated in which the spurious links are omitted.
\subsection{Data, preprocessing, and parcellation}
In this study, we utilized task-based fMRI data obtained from the HCP. Specifically, we focused on the movie-watching task data of 172 participants, which is part of the HCP's tasks designed to elicit a broad range of neural activity. Preprocessing of the fMRI data was conducted using established pipelines provided by the HCP. For this study, the brain was divided into 100 distinct regions using the Schaefer parcellation \cite{schaefer2018local} scheme, and 16 subcortical regions are added based on the Destrieux parcellation \cite{destrieux2010automatic} to ensure causal sufficiency.

\section{Experiments}
\subsection{Experiment design}
The relationship between the visual and attention networks during movie-watching tasks is a subject of interest within neuroscience, given that both networks are highly engaged in processing the complex, dynamic stimuli presented in films. The visual and attention networks comprise 13 and 27 nodes, respectively. We explore the interaction between these networks by identifying nodes from the attention (visual) network that act as confounding variables for at least two nodes in the visual (attention) network to demonstrate our method's performance and discuss the consistency of our results with studies such as \cite{ptak2012frontoparietal}.
\subsection{Results}
Table 1 shows the candidate confounder variables of the visual network pairs within the attention network derived from steps 1 and 2 of our proposed algorithm.

\begin{table}[]
\centering
\label{table:11}
\caption{The candidate confounders in the Attention network and the corresponding variables in the Visual network. V1, V2, V3, V4, V5, and V6 denote LH\_VisCent\_ExStr\_2, LH\_VisCent\_ExStr\_3, RH\_VisCent\_ExStr\_1, RH\_VisCent\_ExStr\_3, RH\_VisPeri\_ExStrInf\_1, LH\_VisPeri\_StriCal\_1, respectively.}
\begin{tabular}{c|c}
\hline
\multicolumn{1}{|c|}{\begin{tabular}[c]{@{}c@{}}\textbf{Confounding candidates}\\ \textbf{in the Attention network}\end{tabular}} & \multicolumn{1}{c|}{\textbf{Visual network pairs}} \\ \hline
LH\_DorsAttnA\_TempOcc\_1                                                                                                  & (V1, V2), (V1, V3), (V1, V4)              \\ \hline
LH\_DorsAttnA\_SPL\_1                                                                                                      & (V4, V5)                                  \\ \hline
RH\_DorsAttnA\_TempOcc\_1                                                                                                  & (V1, V2), (V1, V4)                        \\ \hline
RH\_DorsAttnA\_SPL\_1                                                                                                      & (V6, V4), (V4, V5)                        \\ \hline
\end{tabular}
\end{table}

Table 2 shows the candidate confounder variables of the attention network pairs within the visual network derived from the same process as in Table 1.

\begin{figure}
\includegraphics[width=\textwidth]{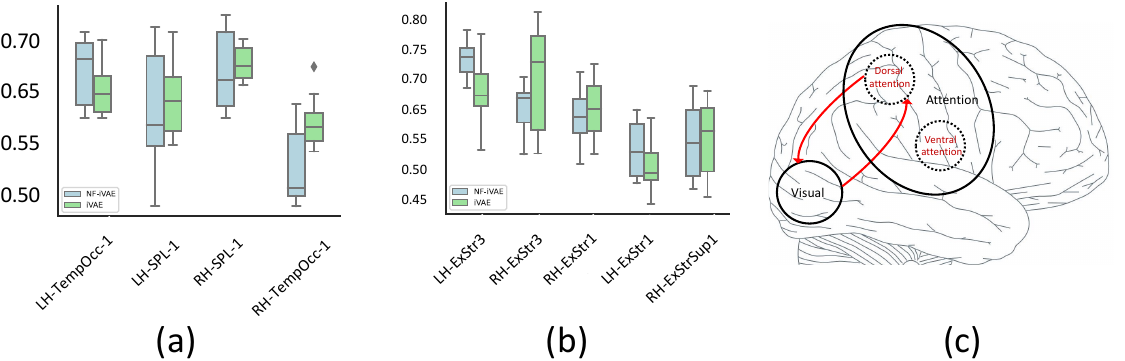}
\caption{ The CCI values for identified regions based on NF-iVAE and iVAE. a. results for the case that attention is confounder b. results for the case that visual is confounder c. the relation between dorsal, ventral and visual regions }\label{fig3_0}
\end{figure}

\begin{table}[]
\centering
\label{table:21}
\caption{The candidate confounders in the Visual network and the corresponding variables in the Attention network. A1, A2, A3, A4, A5, A6, A7, and A8 denote LH\_DorsAttnA\_ParOcc\_1, RH\_DorsAttnA\_SPL\_1, LH\_DorsAttnA\_TempOcc\_1, LH\_DorsAttnB\_FEF\_1, RH\_DorsAttnB\_FEF\_1, RH\_DorsAttnA\_TempOcc\_1, RH\_DorsAttnA\_ParOcc\_1, LH\_DorsAttnA\_SPL\_1, respectively.}
\begin{tabular}{c|c}
\hline
\multicolumn{1}{|c|}{\begin{tabular}[c]{@{}c@{}}\textbf{Confounding candidates}\\ \textbf{in the Visual network}\end{tabular}} & \multicolumn{1}{c|}{\textbf{Attention network pairs}}                                                                                         \\ \hline
LH\_VisCent\_ExStr\_1                                                                                                 & (A1, A2)                                                                                                                             \\ \hline
LH\_VisCent\_ExStr\_3                                                                                                 & \begin{tabular}[c]{@{}c@{}}(A3,A4), (A3, A5), (A1, A8), (A1, A2), (A8, A6), \\  (A8, A7), (A4, A6), (A6, A2), (A6, A5), (A7, A2)\end{tabular} \\ \hline
RH\_VisCent\_ExStr\_1                                                                                                 & (A1, A8), (A1, A2)                                                                                                                   \\ \hline
RH\_VisCent\_ExStr\_3                                                                                                 & \begin{tabular}[c]{@{}c@{}}(A1, A8), (A1, A2), (A8, A6), (A8, A7), (A6, A2),\\  (A7, A2)\end{tabular}                                \\ \hline
RH\_VisCent\_ExStrSup\_1                                                                                              & (A1, A2)                                                                                                                             \\ \hline
\end{tabular}
\end{table}
To identify the confounders among the candidates, by considering the index of the subjects $J$ as the conditioning variable, we apply NF-iVAE model and estimate the CCI between the latents sampled from the model and the confounding variable candidates.
These two Tables demonstrate that mostly the confounding candidates with a higher number of KS-test rejections, acquire higher CCI. 
In Figs \ref{fig3_0}.a and b, 
we include the algorithm's results using the NF-iVAE and iVAE instead of the NF-iVEA. 
According to these results, NF-iVAE acquires noticeably higher CCI than iVAE, since the NF-iVAE can account for the cases where there is dependence between latent variables. Based on the computed results, the identified confounding variables in attention networks are among the Dorsal regions. Additionally, the regions that the visual networks are confounding are in dorsal regions as well. 


Fig \ref{fig3_0}.c, shows the interaction between attention and visual networks.
These findings closely mirror previous research such as \cite{ptak2012frontoparietal}, in which, it is shown that the regions in the dorsal and ventral attention have influenced each other, however, only the Dorsal regions have causal interactions with visual networks. Additionally,  similar to the same study's results, the visual regions also affect the dorsal attention. It is worth noting that in the context of movie-watching, one might hypothesize that the visual network's processing of visual stimuli causally influences the attention network's allocation of attentional resources. However, the reverse could also be true, where the attention network's prioritization influences how visual information is processed.

\begin{figure}
\includegraphics[width=\textwidth]{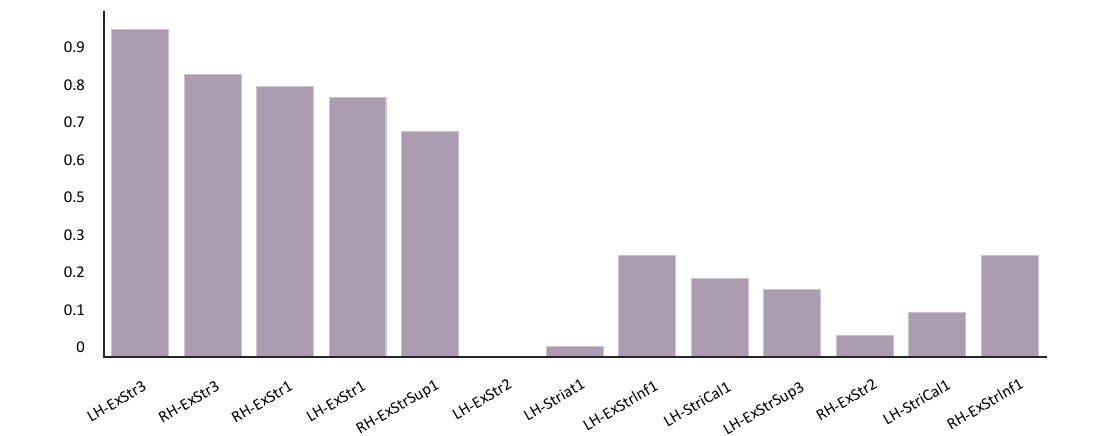}
\caption{ Probability of Visual Cortex Nodes Ranked Among Top 5 by CCI Across 30 Runs. The y-axis denotes the frequency of inclusion in the top 5 rankings, and the x-axis lists the nodes. }\label{fig3}
\end{figure}

To demonstrate the reliability of the identification algorithm, in Fig \ref{fig3}, we present the probability at which each node of the visual network was ranked among the top 5 for having the highest CCI across all 30 runs. In each, the NF-iVAE is initialized with a unique random seed to ensure the diversity of model conditions. In each run, we estimated the CCI  between the latent variables extracted by the NF-iVAE and the original nodes of the visual cortex. This selection process highlighted the nodes that are most effectively captured by the NF-iVAE, reflecting the approach's capacity to consistently identify the most relevant and representative nodes across different initial conditions.
This probability distribution elucidates our method's ability to reliably identify the essential confounding nodes within the visual cortex, underscoring its effectiveness in identifying confounding nodes from proposed candidates.

\section{Conclusion}
This study introduced a novel algorithmic approach for identifying essential exogenous nodes to uphold the assumption of causal sufficiency in the analysis of brain networks. Through an experimental setup involving the visual and attention networks during movie-watching tasks, our method has demonstrated a significant capacity to identify confounding variables that align with neuroscientific studies and highlight the reliability of the proposed method. Moreover, the consistency of iVAE, and NF-iVAE for different random initialization along with high CCI for the proposed candidates driven by independence tests, bolsters the validity of our findings from a neuroscientific perspective.

\appendix

\title{Supplementary for “Algorithmic Identification of Essential
Exogenous Nodes for Causal Sufficiency in Brain Networks"}
%
%
\author{}
\institute{}
%
\authorrunning{A. Bagheri et al.}
%
%
\maketitle              
%


\section{Definitions and preliminaries required for the proposed method} 

To present and discuss the design of the proposed identification method, we first present definitions and preliminaries, including the definition of causal sufficiency, the PC algorithm, an (un)conditional independence test, and the NF-iVAE. Then the algorithm for identifying essential exogenous nodes for causal efficiency is presented. Finally, we describe the data used and the preprocessing steps.
In this paper, $v \in V$ shows all the variables. $z \in Z$ are the variables in the network under the study. $s \in S$ is variables out of the networks under study, i.e., candidate confounders and $c \in C$ are the confounders. $j \in J$ denotes the index of the subjects in the set $J$. 
\subsection{Definitions and preliminaries}

\subsubsection{Causal sufficiency}\
A general definition of causal sufficiency assumption is presented in \cite{beckers2021causal} as follows,

\begin{definition}[A general definition of causal sufficiency]
The set of variables in $S\subseteq V$ is causally sufficient for all the variables in $Z\subseteq V$, if, for every pair of variable $z_1$ and $z_2$, all the common causes or at least the ancestor of common causes are in $S$, i.e., $C\subseteq S\subseteq V$.
\end{definition}

\subsubsection{Peter-Clark algorithm}\

The Peter-Clark (PC) algorithm \cite{spirtes2000causation} provides a search architecture under i.i.d. sampling. 
This algorithm performs as follows, 
1- PC starts with a complete undirected graph. 2- It removes edges between variables that are unconditionally independent of each other. This step eliminates edges representing direct relationships between variables that are not supported by the data. 3- For each pair of variables with an edge between them, and for each variable connected to either of them: If the pair of variables remains independent when conditioned on the connecting variable, the algorithm removes the edge between them.
This step eliminates edges where the relationship between variables can be explained by a third variable.
4- If the pair of variables remains independent when conditioned on both pairs of connecting variables, remove the edge between them.



\subsubsection{Unconditional and conditional independence test}
\begin{lemma}[Independence test]\cite{zhang2011kernel}
Under the null hypothesis that \( v_1 \) and \( v_2 \) are statistically independent, the statistic
\begin{equation}
    \hat{T}_{UI}= \frac{1}{n} \text{Tr}(\hat{K}_{v_1} \hat{K}_{v_2})
\end{equation}
has the same asymptotic distribution as
\begin{equation}
    \hat{T}_{UI} = \frac{1}{n^2} \sum_{i,j=1}^{n} \lambda_{{v_1},i} \lambda_{{v_2},j} \zeta_{ij}^2,
\end{equation}
i.e., \( T_{UI} \overset{d}{=} \hat{T}_{UI} \) as \( n \to \infty \). In this lemma, $\hat{K}_{v_1}$ and $\hat{K}_{v_2}$ are the corresponding centralized kernel matrices. Given the samples $v_1$ and $v_2$, $\lambda_{v_1,i}$ and  $\lambda_{v_2,j}$ are the eigenvalues of $\hat{K}_{v_1}$ and $\hat{K}_{v_2}$, respectively. \( \zeta_ij \) are i.i.d. standard Gaussian variables (i.e., \( zeta_{ij}^2 \) are i.i.d. $ \chi_1^2$-distributed variables).

\end{lemma}

\begin{lemma}[Conditional independence test]\cite{zhang2011kernel}
Under the null hypothesis \( H_0 \) ($v_1$ and $v_2$ are conditionally independent given z), we have that the statistic
\begin{equation}
    T_{CI} = \frac{1}{n} \text{Tr}(\hat{K}_{v_1|z} \hat{K}_{v_2|z})
\end{equation}
has the same asymptotic distribution as
\begin{equation}
    \hat{T}_{CI} = \frac{1}{n} \sum_{k=1}^{n^2} \hat{\lambda}_k \cdot \zeta_k^2,
\end{equation}
where \( \hat{\lambda}_k \) are eigenvalues of \( \hat{w}\hat{w}^T \) and \( \hat{w} = [\hat{w}_1, \ldots, \hat{w}_n] \), with the vector \( \hat{w}_t \) obtained by stacking
$ M_t =[ 
\psi_{v_1|z,1}(\hat{v_1}_t),
...,
\psi_{v_2|z,n}(\hat{v_2}_t)]
[\psi_{v_2|z,1}(\hat{v_2}_t),
...,
\psi_{v_2|z,n}(\hat{v_2}_t)]^T. $
The eigenvalue decompositions (EVD) of the centralized kernel matrices are 
$ \tilde{K}_{v_1} = V_{v_1} A_{v_1} V_{v_1}^T $ 
and
$ \tilde{K}_{v_2} = V_{v_2} A_{v_2} V_{v_2}^T $,
where $ A_{v_1}$ and $ A_{v_2} $ are the diagonal matrices containing the non-negative eigenvalues $ \lambda_{v_2,i} $ and $ \lambda_{y,i} $, respectively. Then, $ \psi_{v_1} = [\psi_1(v_1), \ldots, \psi_n(v_1)] V_{v_1} A_{v_1}^{1/2} $ and $ \phi_y = [\phi_1(v_2), \ldots, \phi_n(v_2)]  V_{v_2} A_{v_2}^{1/2} $. I.e., $ \psi_i(v_1) = \sqrt{\lambda_{v_1,i}} V_{v_1,i} $, where $ V_{v_1,i} $ denotes the $i $th eigenvector of $ K_{v_1} $.

\end{lemma}




\subsubsection{NF-iVAE}\

NF-iVAE can be used to identify latent variables for a model in non-stationary environments \cite{lu2021invariant}, contrary to the VAE's that are frequently used for the identification of latent confounding variables, \cite{tran2015variational,harada2022infocevae}. NF-iVAE posits a more general setting than iVAE \cite{khemakhem2020variational} for which the dependence between latent variables is considered in Prior the distribution. It assumes the conditional prior $p(S|J)$ belongs to a general exponential family formalized as follows: $p(S|J)$ belongs to a general exponential family with a parameter vector given by an arbitrary function $U$  and sufficient statistics $T(S) = [T_f(S)^T, T_{NN}(S)^T]^T $ given by the concatenation of: a) the sufficient statistics $T_f(S) = [T_1(s_1)^T, \ldots, T_n(s_n)^T]^T$ of a factorized exponential family, where $s_i \in S$ and for all the $T_i(s_i)$ have dimension larger or equal to $2$, and b) the output $T_{NN}(S)$ of a neural network with ReLU activations.
The resulting density function is thus given by
\begin{equation}
p_{T,\lambda}(S|J) = \frac{Q(S)}{C'(J)} \exp\left[T(S)^T \lambda(J)\right],
\end{equation}
where $Q$ is the base measure and $C'$ is the normalizing constant.

\subsection{Identification of Essential exogenous nodes for causal sufficiency}


Our proposed identification algorithm runs in three main steps:

\begin{enumerate}
     \item Our approach, similar to the PC algorithm, initiates by conducting unconditional and conditional independence tests between all pairs of regions, $Z$. Complementary to the PC algorithm, it then carries out conditional independence tests on these same pairs conditioned on regions from other networks, $S$ is the number of possible confounders. This procedure is uniformly applied across all participants.
     \item By comparing the distribution of conditional and unconditional independence tests of each pair for all subjects with the Kolmogorov-Smirnov (KS) test, a subset of candidate confounding variables are distinguished. These candidates can represent both a confounder and a mediator.

     \item Measuring the Correlation coefficient index (CCI) between the latent variables generated by NF-iVAE and the real values of candidates proposed in the previous step, we identify the nodes that are confounding variables of the network under study.

\end{enumerate}

It is important to note that while NF-iVAE can identify all confounding variables, it necessitates the fulfillment of certain stringent assumptions, such as the condition (iv) in Theorem 1 \cite{lu2021invariant}, which stipulates the existence of $k + 1$ distinct points $\{U^0, U^1, \ldots, U^k\}$ where the matrix $L = [\lambda(U)^1 - \lambda(U)^0, \ldots, \lambda(U)^k - \lambda(U)^0]$ of size $k \times k$ is invertible, with $k$ representing the dimension of $T$. Furthermore, in NF-iVAE method, the number of confounding variables and topological order has to be known. This imposes limitations in scenarios where the number of nodes in the target network, whose causal structure is to be determined, is fewer than the number of confounding variables. Conversely, relying on candidates discovered in the first two steps offers an alternative approach for identifying potential confounding variables in situations where the availability of distinct points $\{U^0, U^1, \ldots, U^k\}$ and knowledge of the topological order of nodes are insufficient. This results in more effectiveness even when the number of confounding variables exceeds the number of nodes for which causal structure identification is sought as we only apply the NF-iVAE on the nodes of interest.

\end{document}